\documentclass[letterpaper, 10 pt, conference]{ieeeconf}  

\IEEEoverridecommandlockouts                              

\usepackage{cite}
\usepackage{amsmath,amssymb,amsfonts}
\usepackage{algorithmic}
\usepackage{graphicx}
\usepackage{textcomp}
\usepackage{xcolor}

\usepackage{bm}
\usepackage{mathtools}
\usepackage{booktabs}
\usepackage{subcaption}

\title{\LARGE \bf
Tunable Virtual IMU Frame by Weighted Averaging of Multiple Non-Collocated IMUs
}

\author{Yizhou Gao and Timothy D. Barfoot
\thanks{Yizhou Gao and Timothy D. Barfoot are with the University of Toronto Robotics Institute, Toronto, Ontario, Canada (email: yizhou.gao@robotics.utias.utoronto.ca; tim.barfoot@utoronto.ca).}
}

\begin{document}

\maketitle
\thispagestyle{empty}
\pagestyle{empty}

\begin{abstract}
We present a new method to combine several rigidly connected but physically separated IMUs through a weighted average into a single virtual IMU (VIMU). This has the benefits of (i) reducing process noise through averaging, and (ii) allowing for tuning the location of the VIMU. The VIMU can be placed to be coincident with, for example, a camera frame or GNSS frame, thereby offering a quality-of-life improvement for users. Specifically, our VIMU removes the need to consider any lever-arm terms in the propagation model.  We also present a quadratic programming method for selecting the weights to minimize the noise of the VIMU while still selecting the placement of its reference frame. We tested our method in simulation and validated it on a real dataset. The results show that our averaging technique works for IMUs with large separation and performance gain is observed in both the simulation and the real experiment compared to using only a single IMU.
\end{abstract}

\section{Introduction}

An inertial measurement unit combining a gyroscope and accelerometer that provides measurements of angular velocity and linear acceleration has wide applications in robotics. By integrating the measurements, one can track the relative orientation and position of the system to a certain degree of accuracy over a reasonable time frame \cite{tang2022_preintegration}. However, due to the interoceptive nature of an IMU, it is often paired with additional sensors such as a camera, lidar, or GNSS to reduce drift \cite{alaba2024_gps_and_imu, shan2020_liosam, he2020_visual_imu_and_gps}.

There has also been great interest in exploiting multiple IMUs to improve the accuracy of state estimators. For instance, there have been hundreds of papers on inertial sensor arrays, as summarized in \cite{inertial_sensor_array}. There are also many works detailing methods of fusing multiple IMU measurements together \cite{patel2022_multi-imu}. Methods for calibrating multiple IMUs have also been proposed \cite{Multi_vins, MI-Calib}.  To improve robustness, there have been studies using redundant IMUs for fault detection and to build fault-tolerant systems. For example, to detect a measurement error, \cite{Sturza1988_redundant} proposed a parity vector based on hypothesis testing. The optimal geometric configuration of multiple redundant IMUs is also explored in \cite{Colomina2004REDUNDANTIF, guerrier2009, xue2023}.

In this article, we will show that through a simple weighted averaging of multiple non-collocated IMUs, we not only offer the benefit of reduced noise and bias drift, but also the freedom to choose an arbitrary VIMU frame of reference (as shown in Figure 1) such that all of the IMUs can be treated as one.  Placing the VIMU frame to be coincident with a camera, for example, could be very convenient by avoiding the need to consider lever-arm effects.

In Section \ref{related_work}, we will compare different approaches for fusing multiple IMUs, and in Section \ref{methodology} we will discuss the main idea of how to average multiple non-collocated IMUs. Finally, in Sections \ref{simulation} and \ref{real_experiment}, we validate our findings through experiments with simulation and a real dataset.

\begin{figure}
    \centering
    \includegraphics[width=0.5\linewidth]{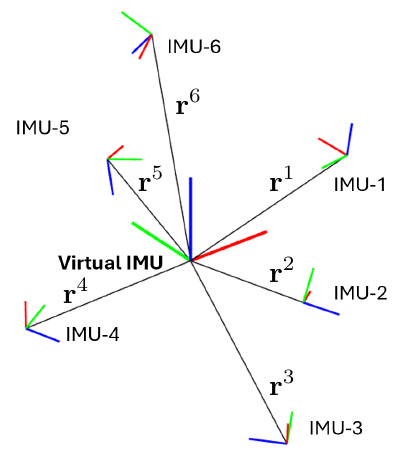}
    \caption{We propose a virtual IMU through a weighted average of the outputs of several physically separate IMUs. By tuning the weights $w^j$ while ensuring that $\sum_j w^j \mathbf{r}^j = \mathbf{0}$, our VIMU's propagation model has no lever-arm terms (acceleration from rotation), its location can be arbitrarily selected (in the convex hull of the individual IMUs), and its measurement noise can be minimized.}
    \label{fig:vimu_frame}
    \vspace*{-0.2in}
\end{figure}

\section{Related Work}\label{related_work}

In this section, we will introduce some prior work on fusing multiple IMUs for use in an estimation pipeline.

\subsection{Sensor Level Fusion - Virtual IMU}\label{VIMU}

One approach is to perform the fusion of multiple IMUs into a VIMU before using the measurements in an inertial navigation system (INS). For example, \cite{jafari2014_PEM} proposed a Prediction Error Minimization method to model the noise of IMUs and use a Kalman filter to estimate the optimal process signal at the VIMU frame. \cite{xue2023} proposed using a Kalman filter to estimate the angular velocity measured by multiple non-orthogonal gyroscopes. Instead of using a Kalman filter, an easier approach is to directly average multiple gyroscopes after alignment to a common virtual IMU frame, as discussed in \cite{waegli2008, patel2022_multi-imu, Colomina2004REDUNDANTIF}.

Unlike averaging gyroscopes, averaging multiple accelerometers is not trivial due to the additional acceleration terms arising from rotation (lever-arm effect), unless they are closely placed and the separations between them are ignored. \cite{zhang2020} addressed this by mapping measurements from all IMUs onto the VIMU frame with a least-squares estimator. Due to angular acceleration not being measured by the IMUs, it was marginalized out in the derivation. The resulting equation for fusing multiple accelerometers is complicated and carries an extra term to account for the lever-arm effect. Closest to our work, \cite{Fu2023} performs averaging of multiple IMUs but without considering the acceleration bias. In Section \ref{methodology}, we will show a simpler way to average multiple accelerometers with large separations and without approximation. Instead of averaging, \cite{best_axis_composition} proposed composing the VIMU by choosing the best three axes from all of the gyroscopes and accelerometers. However, this approach limits the noise reduction potential as it effectively uses only a single IMU per axis.

\subsection{State Augmentation with Multiple IMUs}\label{augmented}

Typically, a state estimation problem with an IMU involves tracking the orientation, linear velocity, and linear acceleration of the vehicle (commonly defined at the IMU). With multiple IMUs, additional state variables such as linear acceleration, angular velocity, and angular acceleration are added. This allows using the full mechanization model of the accelerometer as the measurement update in the Kalman filter \cite{Bancroft2011DataFA, Beaudoin2018_satelite}. On the contrary, our approach only involves averaging the measurements, thus keeping the computational overhead minimal.

\subsection{Stacked IMU States with Geometric Constraints}\label{constraint}

Another approach is to track the state of each IMU individually and associate them with geometric constraints \cite{waegli2008, Beaudoin2018_satelite}. Using multiple IMUs and cameras, \cite{Eckenhoff2021_MIMC-VINS} proposed tracking all stacked states of base and auxiliary IMUs using MSCKF \cite{Anastasios2007_MSCKF}. Assuming the extrinsic calibrations of all IMUs are known and fixed, relative geometric constraints are imposed between states of different IMUs in the form of measurement updates, in addition to camera observations. Our approach allows fusing the information from multiple IMUs without the need to introduce extra geometric constraints in the state estimation pipeline.

\subsection{Stacked IMU States with Calibration Parameters}\label{constraint}

Unlike methods in \ref{constraint}, \cite{yang2024MVIS} proposes to estimate sensor intrinsic/extrinsic jointly along with the states of each IMU and camera under an optimization framework. Due to the number of states variables, one can expect this approach to be computationally heavy. Also, careful observability analysis is required as detailed in the paper.

\subsection{Distributed INS}\label{distributed}

The last common approach is to have an individual state estimator for each IMU, including local propagation and correction. These separate estimates are then fused by a master filter \cite{Bancroft2011DataFA, patel2022_multi-imu, jadid2019} or simply weighted-averaged \cite{Wu2023, grover2024}. Our approach avoids the need to run several state estimators, which can be expensive.

\subsection{Summary}

As presented, multiple approaches have been proposed. \ref{constraint} is computationally expensive if the number of IMUs is large, as it has many states all in one Kalman filter. The averaging approach in \ref{VIMU} is the simplest to implement and does not significantly alter the existing single-IMU framework. However, due to the lever-arm effect, averaging multiple accelerometers is a challenge. We will show a new way to simply average the accelerometers despite the separation and give a tunable VIMU frame as a result.

\section{Preliminaries}

\subsection{Definitions}

We define the following frames of reference: 1) inertial frame $\bm{\mathcal{F}}_i$, 2) sensor frame $\bm{\mathcal{F}}_s$ as the local frame of reference at each sensor, and 3) VIMU frame or vehicle frame $\bm{\mathcal{F}}_v$ for tracking vehicle pose in the state estimation pipeline.\footnote{Estimation is often performed in the IMU frame to remove the need to deal with lever-arm terms in the propagation model.}


For any point, $P$, in the inertial frame $\bm{\mathcal{F}}_i$, we denote it as $\textbf{r}_i^{pi}$. For example, The vehicle position in the inertial frame can be expressed as $\textbf{r}_i^{vi}$.

When referring to the pose of a vehicle, it is the tuple $\{\textbf{C}_{iv}, \textbf{r}_i^{vi}\}$, which forms the transformation matrix as follows:
$$
\textbf{T}_{iv} = \left[\begin{matrix}
    \textbf{C}_{iv} & \textbf{r}_i^{vi} \\
    \textbf{0} & 1
\end{matrix}\right].
$$

\noindent The pose can be used to transform a point $\textbf{r}_v^{pv}$ expressed in the vehicle frame $\bm{\mathcal{F}}_v$ into the inertial frame $\bm{\mathcal{F}}_i$ as $\textbf{r}_i^{pi}$:
$$
\left[\begin{matrix} \textbf{r}_i^{pi} \\ 1 \end{matrix}\right]
    = \textbf{T}_{iv} \left[\begin{matrix} \textbf{r}_v^{pv} \\ 1 \end{matrix}\right]
\quad \text{or} \quad
\textbf{r}_i^{pi} = \textbf{C}_{iv} \textbf{r}_v^{pv} + \textbf{r}_i^{vi}.
$$

\noindent In this document, $^\wedge$ is specifically used for rendering an element of $\mathbb{R}^3$ as a skew-symmetric matrix:
$$
\bm{\omega}^\wedge = \left[\begin{matrix}
    0 & -\omega_2 & \omega_1 \\
    \omega_2 & 0 & -\omega_3 \\
    -\omega_1 & \omega_3 & 0
\end{matrix}\right].
$$

\noindent If a quantity is in the form $x(t)$, then it is a time-varying quantity, otherwise it is time-invariant.

\subsection{Rigid-Body Kinematics}

The state propagation using IMU measurements follows the rigid-body kinematics in the inertial frame:
\begin{equation}
\begin{split}
    \dot{\textbf{C}}(t) &= \textbf{C}(t) \bm{\omega}(t)^\wedge, \\
    \dot{\textbf{v}}(t) &= \textbf{C}(t) \textbf{a}(t) - \textbf{g}, \\
    \dot{\textbf{p}}(t) &= \textbf{v}(t).
\end{split}
\end{equation}
\noindent Here $\textbf{C}$ is short for rotation $\textbf{C}_{iv}$, $\textbf{v}$ is short for velocity $\textbf{v}_i^{vi}$, and $\textbf{p}$ is for position $\textbf{p}_i^{vi}$. These three form a 9-DOF state for the vehicle.

In addition, $\textbf{g}$ is the acceleration from gravity and $\bm{\omega}$ and $\textbf{a}$ are the true angular velocity and true linear acceleration in the vehicle frame, respectively.

\subsection{IMU Measurement Model}

For the angular velocity measurement from a gyroscope, we have the following measurement model:
\begin{equation}
    \textbf{y}_\omega(t) = \bm{\omega}(t) + \textbf{b}_\omega(t) + \textbf{n}_\omega(t),
\end{equation}
\noindent where $\textbf{b}_\omega$ is the bias of the gyroscope measurement and $\textbf{n}_\omega$ is the additive white noise of the measurement. Similarly, we have the following measurement model for an accelerometer coincident with the vehicle frame:
\begin{equation}
    \textbf{y}_a(t) = \textbf{a}(t) + \textbf{b}_a(t) + \textbf{n}_a(t),
\end{equation}
\noindent where $\textbf{b}_a$ is the bias of the accelerometer and $\textbf{n}_a$ is the noise of the acceleration measurement.

Bias drifts of the IMU are modeled as a random walk with the bias drift rate following a Gaussian process:
\begin{equation}
\begin{split}
    \dot{\textbf{b}}_\omega(t) &= \textbf{w}_\omega(t), \\
    \dot{\textbf{b}}_a(t) &= \textbf{w}_a(t).
\end{split}
\end{equation}
\noindent The biases are estimated as part of the state vector and subtracted from the raw measurement to obtain the `unbiased' measurement for propagation:
\begin{equation}
\begin{split}
    \hat{\bm{\omega}}(t) = \textbf{y}_\omega(t) - \hat{\textbf{b}}_\omega(t), \\
    \hat{\textbf{a}}(t)  = \textbf{y}_a(t) - \hat{\textbf{b}}_a(t).
\end{split}
\end{equation}

We assume that we have several IMUs (index $j$) whose extrinsic calibrations, $\{ \mathbf{C}^j, \mathbf{r}^j \}$, are {\em fixed} and {\em known} relative to our desired VIMU frame. For IMU $j$, measurements are in its local frame. For the gyroscope measurement, we have
\begin{equation}
    \textbf{y}^j_{\omega}(t) = \textbf{C}^{j^T} \bm{\omega}(t) + \textbf{b}^j_{\omega}(t) + \textbf{n}^j_{\omega}(t).
\end{equation}

\noindent For the accelerometer, we have additional terms in the acceleration measurement due to the lever-arm effect:
\begin{equation}
\begin{split}
    \textbf{y}^j_{a}(t) = \textbf{C}^{j^T} \left( \textbf{a}(t) + \bm{\omega}(t)^\wedge \bm{\omega}(t)^\wedge \textbf{r}^j + \bm{\alpha}(t)^\wedge \textbf{r}^j \right) \\
    + \; \textbf{b}^j_{a}(t) + \textbf{n}^j_{a}(t).
\end{split}
\end{equation}
where $\bm{\alpha}(t)$ is angular acceleration.


\section{Methodology}\label{methodology}

\subsection{Averaging Multiple Gyroscopes}

First, we align the gyroscope measurements with the vehicle frame:
\begin{equation}
    \textbf{C}^{j} \textbf{y}^j_{\omega}(t) = \bm{\omega}(t) + \textbf{C}^{j} \textbf{b}^j_{\omega}(t) + \textbf{C}^{j} \textbf{n}^j_{\omega}(t).
\end{equation}
\noindent The weighted average of all gyroscopes is simply
\begin{equation}
\begin{split}
    \bar{\textbf{y}}_\omega(t) &= \sum_j{w^j \textbf{C}^{j} \textbf{y}^j_{\omega}(t)} \\
    &= \sum_j{w^j \left( \bm{\omega}(t) + \textbf{C}^{j} \textbf{b}^j_{\omega}(t) + \textbf{C}^{j} \textbf{n}^j_{\omega}(t) \right)} \\
    &= \bm{\omega}(t) + \sum_j{w^j \textbf{C}^{j} \textbf{b}^j_{\omega}(t)} + \sum_j{w^j \textbf{C}^{j} \textbf{n}^j_{\omega}(t)}
\end{split}
\end{equation}
\noindent where $w^j$ are the yet-to-be-determined weights with $\sum_j{w^j} = 1$. We see that the averaged measurement can be summarized as
\begin{equation}
    \bar{\textbf{y}}_\omega(t) = \bm{\omega}(t) + \bar{\textbf{b}}_\omega(t) + \bar{\textbf{n}}_\omega(t),
\end{equation}
\noindent where $\bar{\textbf{b}}_\omega(t)$ and $\bar{\textbf{n}}_\omega(t)$ are the combined biases and noises:
\begin{equation}
    \begin{split}
        \bar{\textbf{b}}_\omega(t) &= \sum_j{w^j \textbf{C}^{j} \textbf{b}^j_{\omega}(t)}, \\
        \bar{\textbf{n}}_\omega(t) &= \sum_j{w^j \textbf{C}^{j} \textbf{n}^j_{\omega}(t)}.
    \end{split}
\end{equation}

\subsection{Averaging Multiple Accelerometers}

Aligning all accelerometer measurements to the vehicle frame and summing over all accelerometers, we have

\begin{equation}
\begin{split}
    \bar{\textbf{y}}_a(t) &= \sum_j{w^j \textbf{C}^{j} \textbf{y}^j_{a}(t)} \\
    &= \textbf{a}(t) + \sum_j{w^j \textbf{C}^{j} \textbf{b}^j_{a}(t)} + \sum_j{w^j \textbf{C}^{j} \textbf{n}^j_{a}(t)} \\
    &\quad + \sum_j{w^j \bm{\omega}(t)^\wedge \bm{\omega}(t)^\wedge \textbf{r}^j} + \sum_j{w^j \bm{\alpha}(t)^\wedge \textbf{r}^j}.
\end{split}
\end{equation}
\noindent Using the fact that $^\wedge$ is linear, we have
\begin{equation}\label{accel_avg_final}
\begin{split}
    \bar{\textbf{y}}_a(t) 
    &= \textbf{a}(t) + \sum_j{w^j \textbf{C}^{j} \textbf{b}^j_{a}(t)} + \sum_j{w^j \textbf{C}^{j} \textbf{n}^j_{a}(t)} \\
    &\quad + \; \bm{\omega}(t)^\wedge \bm{\omega}(t)^\wedge \left( \sum_j{w^j \textbf{r}^j} \right) + \bm{\alpha}(t)^\wedge \left( \sum_j{w^j\textbf{r}^j} \right).
\end{split}
\end{equation}
\noindent We see that if $\sum_j{w^j \textbf{r}^j} = \textbf{0}$, then we can {\em eliminate} the last two terms from the lever-arm effect. This simplifies (\ref{accel_avg_final}) to
\begin{equation}
    \bar{\textbf{y}}_a(t) = \textbf{a}(t) + \bar{\textbf{b}}_a(t) + \bar{\textbf{n}}_a(t),
\end{equation}
with $\bar{\textbf{b}}_a(t) = \sum_j{w^j \textbf{C}^{j} \textbf{b}^j_{a}(t)}$ and $\bar{\textbf{n}}_a(t) = \sum_j{w^j \textbf{C}^{j} \textbf{n}^j_{a}(t)}$.

\subsection{VIMU from Weighted Average}\label{AA}

As we can see, if we choose the VIMU frame to be the same as the vehicle frame\footnote{Note, the vehicle frame can actually be placed arbitrarily, but is a useful construct during the derivation. Hence, we can think of the VIMU and vehicle frames as the same thing moving forward.} and choose the weights of the accelerometers such that $\sum_j{w^j \textbf{r}^j} = \textbf{0}$, then we can ignore the separation of all IMUs and fuse the accelerometer measurements through averaging. The result is a virtual IMU at the vehicle frame without any lever-arm effects.

Since gyroscope measurements are independent of the location, we can choose different weights for averaging gyroscopes and accelerometers.

Because all biases are slowly varying, the combined biases are also slowly varying. Therefore, instead of tracking individual biases, only the combined biases are tracked. When receiving a new set of IMU measurements, we estimate the angular velocity and linear acceleration with the following simple update:
\begin{equation}
\begin{split}
    \hat{\bm{\omega}}(t) = \bar{\textbf{y}}_\omega(t) - \hat{\bar{\textbf{b}}}_\omega(t), \\
    \hat{\textbf{a}}(t)  = \bar{\textbf{y}}_a(t) - \hat{\bar{\textbf{b}}}_a(t).
\end{split}
\end{equation}
The combined measurements are expected to have slower drift and the estimate is expected to be less noisy as well \cite{patel2022_multi-imu}. If we have $n$ identical IMUs and equal weights, the standard deviation of the combined noise or drift rate is expected to be $\bar{\sigma} = \sigma / \sqrt{n}$.

\subsection{Selecting Weights for Minimal Uncertainty}\label{solve_weight_by_noise}

When selecting weights for gyroscopes, or for accelerometers under the condition that the VIMU frame choice is flexible, we can minimize the output noise or bias drift rate. Below, we take the bias drift as an example but similar ideas can be applied to the output noise.



The covariance of the combined drift rate is the weighted sum of the noise covariances of each IMU:
\begin{equation}
    \bar{\bm{\Sigma}} = \sum_j{w^{j^2} \bm{\Sigma}^j}.
\end{equation}
\noindent We can then solve for the weights that minimize the trace of the combined covariance.
If the IMUs have isotropic noise for all three axes, the combined variance simplifies to
\begin{equation}\label{noise_reduction_asym}
    \bar{\sigma}^2 = \sum_j{\left(w^{j} \sigma^{j}\right)^2}.
\end{equation}
\noindent Furthermore, if using $n$ of the same IMUs and equal weighting, the combined variance is simply $\sigma^2/n$.

\subsection{Selecting Weights for VIMU Frame Placement}

Given that $\sum_j{w^j \textbf{r}^j} = \textbf{0}$, with 2 IMUs, we can only place the VIMU frame on the line through the center of the two IMUs. Given 3 IMUs, we can place the VIMU frame on the plane formed by the three IMUs. With 4 non-coplanar IMUs, then in theory, we can place the VIMU frame anywhere. However, placing the VIMU frame outside of the convex hull of the IMUs may increase the output noise. For example, for $n$ identical IMUs, the noise is increased if $\sum_j{(w^j)^2 > 1}$. This is not possible as long as VIMU is within the convex hull ($w^j > 0, \forall j)$.

When we have a target VIMU frame in mind, we need to solve for the required weights. Assuming we have arbitrarily many IMUs, then the problem can be formulated as an optimization problem:
\begin{equation}\label{opt_problem}
\begin{split}
    \min_{w^j}{\frac{1}{2} \sum_j{\left(w^{j} \sigma^{j}\right)^2}} \quad \text{s.t. }
    \begin{cases}
      \sum_j{w^j \textbf{r}^j} = \textbf{0} \\
      \sum_j{w^j} = 1
    \end{cases}
\end{split}.
\end{equation}
We chose the objective to reduce output noise as shown in Section \ref{solve_weight_by_noise}. Since (\ref{opt_problem}) is a quadratic programming problem, it is possible to find a closed-form solution.

The Lagrangian of the problem is $\mathcal{L} = \frac{1}{2}\textbf{w}^T \bm{\Sigma} \textbf{w} + \bm{\lambda}^T \textbf{R}\textbf{w} + \beta \left( 1 - \textbf{1}^T \textbf{w} \right)$ where $\textbf{1}^T = \left[\begin{matrix}1 \cdots 1\end{matrix}\right]$, $\textbf{w}^T = \left[\begin{matrix} w^1 \cdots w^n \end{matrix}\right]$, $\bm{\Sigma} = \text{diag}(\sigma^{1^2}, ..., \sigma^{n^2})$, and $\textbf{R} = \left[\begin{matrix} \textbf{r}^1 \cdots \textbf{r}^n\end{matrix}\right]$ is the stacked matrix of the positions of IMUs in the VIMU frame. This gives the following KKT conditions:
\begin{equation}
\begin{split}
  \bm{\Sigma}\textbf{w} + \textbf{R}^T \bm{\lambda} - \beta \textbf{1} &= \textbf{0}, \\
  \textbf{R} \textbf{w} &= \textbf{0}, \\
  \textbf{1}^T \textbf{w} &= 1.
\end{split}
\end{equation}

\noindent If we left-multiply the first condition by $\textbf{w}^T$, we have
\begin{equation}
\textbf{w}^T\bm{\Sigma}\textbf{w} + \textbf{w}^T\textbf{R}^T\bm{\lambda} - \beta \textbf{w}^T\textbf{1} = 0.
\end{equation}

\noindent Since $\textbf{R}\textbf{w} = \textbf{0}$ and $\textbf{1}^T\textbf{w} = 1$ this implies
\begin{equation}\label{beta}
    \beta = \textbf{w}^T\bm{\Sigma}\textbf{w}.
\end{equation}

\noindent Left-multiplying the first condition by $\bar{\textbf{R}} = \textbf{R}\bm{\Sigma}^{-1}$ gives
\begin{equation}
\underbrace{\textbf{R}\textbf{w}}_{=\textbf{0}} + \bar{\textbf{R}}\textbf{R}^T\bm{\lambda} - \beta\bar{\textbf{R}}\textbf{1} = 0.
\end{equation}

\noindent Here we define $\bar{\textbf{r}} = \bar{\textbf{R}}\textbf{1} = \sum{\textbf{r}^j / \sigma^{j^2}}$ and using the pseudo-inverse of $\bar{\textbf{R}}\textbf{R}^T$, we have
\begin{equation}\label{lambda}
    \bm{\lambda} = \beta \left(\bar{\textbf{R}}\textbf{R}^T\right)^{+} \textbf{r}.
\end{equation}
\noindent Substituting (\ref{beta}) and (\ref{lambda}) into the first condition we have
\begin{equation}
    \hat{\textbf{w}} = \frac{\textbf{w}}{\textbf{w}^T\bm{\Sigma}\textbf{w}} = \bm{\Sigma}^{-1} \left( \textbf{1} - \textbf{R}^T \left(\bar{\textbf{R}}\textbf{R}^T\right)^{+} \bar{\textbf{r}} \right).
\end{equation}

\noindent Lastly, since $\sum_j{w^j} = 1$, we need to normalize the intermediate solution to get the final weights for the accelerometers:
\begin{equation}
    \textbf{w}^* = \frac{\hat{\textbf{w}}}{\textbf{1}^T \hat{\textbf{w}}}.
\end{equation}

\section{Simulation} \label{simulation}

To check the correctness of the frame-selection method through weighted averaging, and to show the performance benefit of the fused VIMU, we first performed an evaluation in simulation.

\subsection{Data Generation}

To generate the simulated data, we used different combinations of sinusoidal functions to generate the ground-truth angular velocity and linear acceleration.
We then integrate them to generate the ground-truth trajectory. We selected MPU6050, a common MEMS IMU in the mechatronics community, as the example model. For each IMU, we simulated its measurements and added noises and biases. Each IMU was sampled at 100 Hz.

\subsection{Implementation}

We also simulated observations of known landmarks from a monocular camera. This turns the estimation problem into a localization problem, such that the error is bounded. For each camera frame, we first sampled random observations in the imaging plane as well as depth (between 2 and 10 m). We performed reprojection and transformed the resulting 3D points to the inertial frame with the ground-truth pose to generate the corresponding landmark positions. Camera observations were simulated at 2 Hz. We selected this low update rate to allow the error to grow large enough to emphasize the benefits of IMU averaging.

For the estimator, we implemented the \textit{left-}Invariant Extended Kalman Filter (IEKF) \cite{IEKF} as it is more accurate compared to standard EKF methods. At the initialization stage, given all IMU poses, we computed the weights that place the VIMU frame at the camera frame. Using the weights, we also calculated the combined noises and bias drift rates. When new IMU measurements arrived, they were weight-averaged and then fed into the estimator as though we had a single IMU. When camera observations arrived, the 2D observations and 3D landmark positions were used for the correction step.

\subsection{Experiment Setup}

\begin{figure}[t]
    \centering
    \begin{subfigure}[b]{0.45\linewidth}
        \centering
        \hspace*{-0.5in}
        \includegraphics[width=5.5cm]{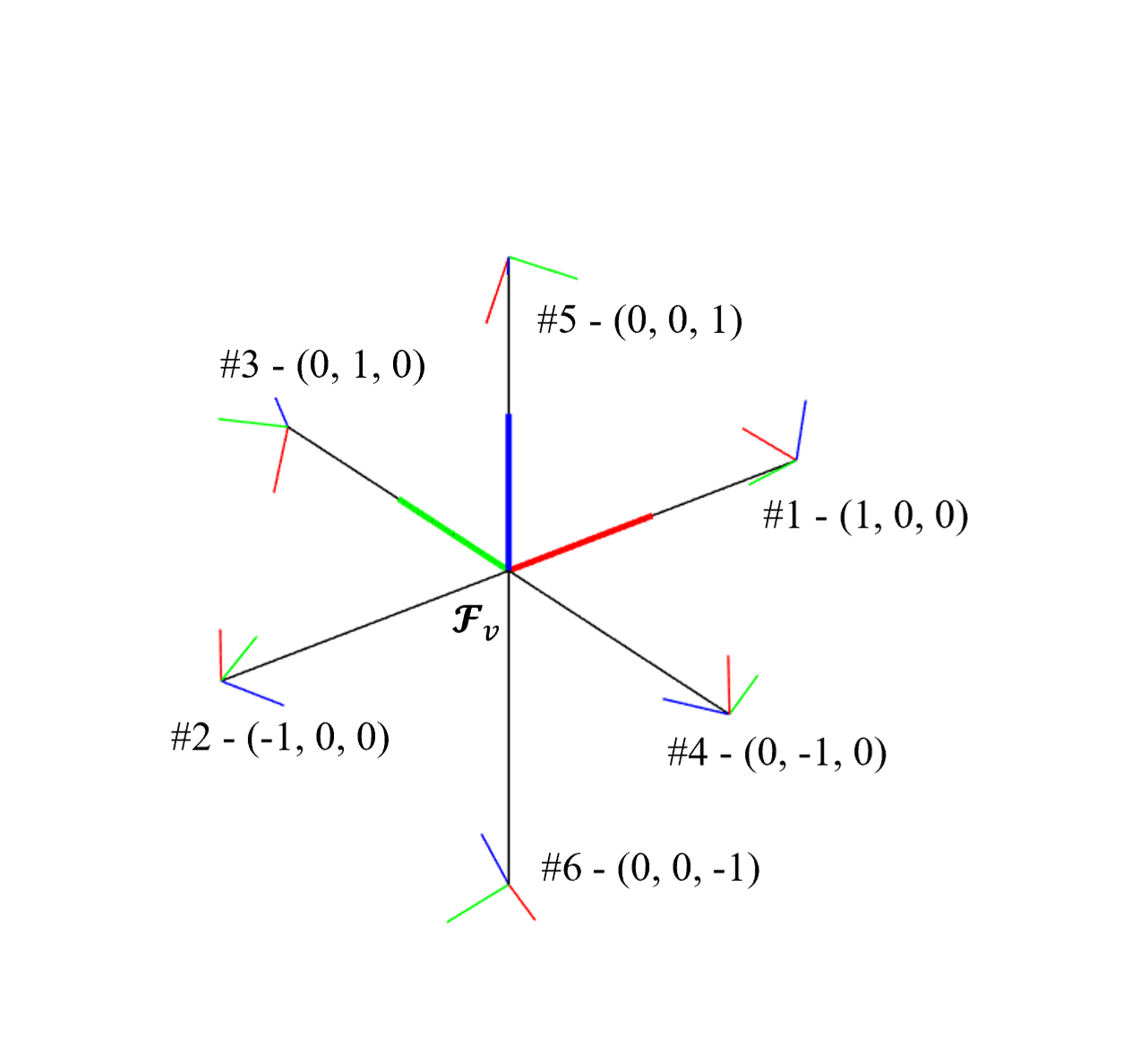}
        \caption{Symmetrical IMUs}
        \label{fig:sym_imu}
    \end{subfigure}
    \begin{subfigure}[b]{0.45\linewidth}
        \centering
        \hspace*{-0.4in}
        \includegraphics[width=5.5cm]{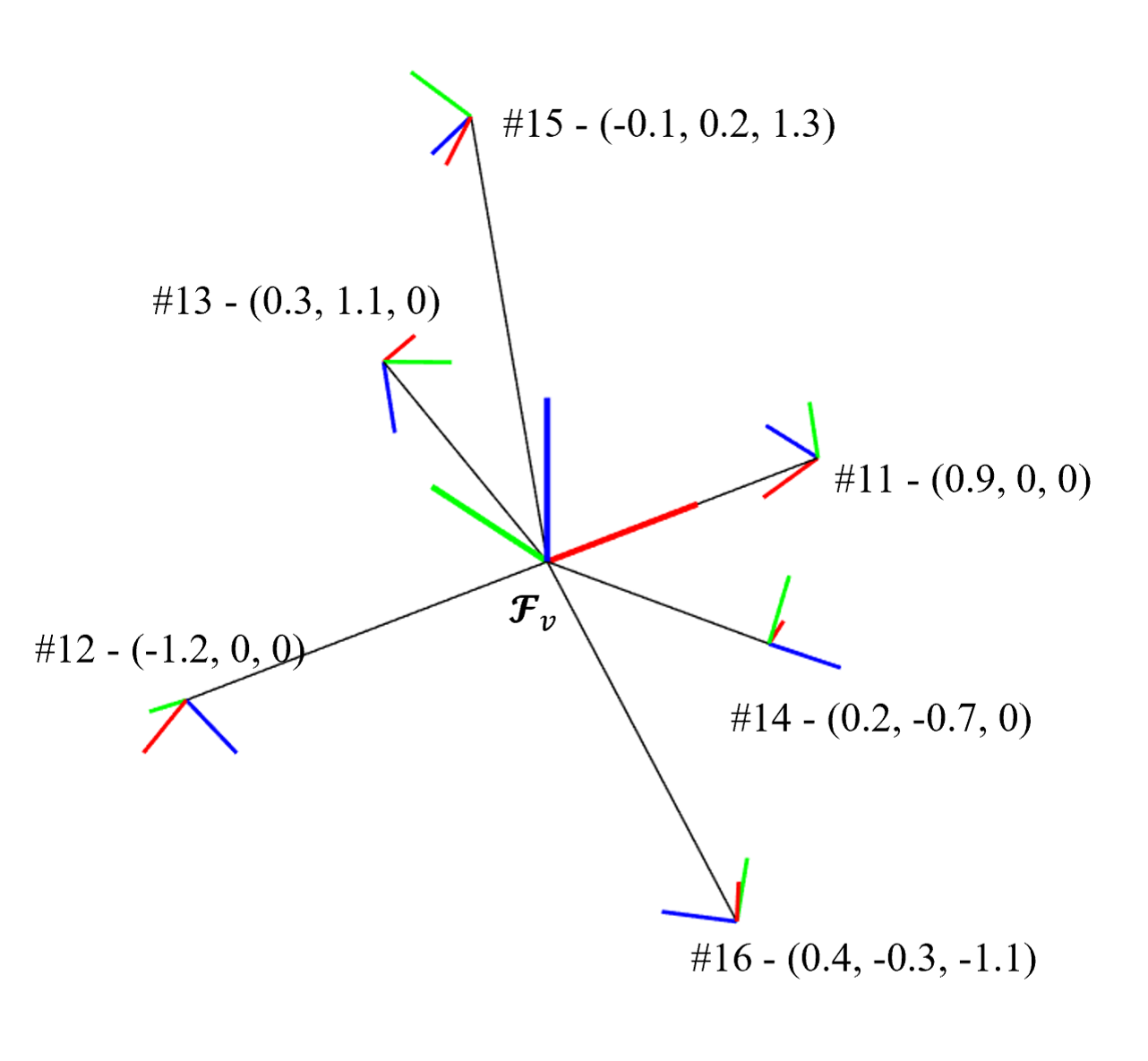}
        \caption{Asymmetric IMUs}
        \label{fig:_asym_imu}
    \end{subfigure}
\caption{Poses of IMUs in simulated configurations. The VIMU frame was chosen to be the same as the camera frame, making the overall estimator easier to implement.}
\label{fig:imu_configurations}
\vspace*{-0.2in}
\end{figure}

For performance, we compared the results of multiple different IMU configurations summarized in Table \ref{tab:imu_config}. Here symmetric configurations have IMUs placed along the principal axes with an offset of 1 m with random orientations. Asymmetric configurations use IMUs perturbed from the symmetric configurations, such that unequal weightings are required to place the VIMU frame at the camera frame. We placed the VIMU frame at the camera frame for all configurations to provide a fair comparison of performance (direct comparison with ground-truth). The poses of all IMUs used are plotted in Figure \ref{fig:imu_configurations}.

\begin{table}[t]
\vspace*{3pt}
\centering
\caption{Simulated IMU Configurations}
\label{tab:imu_config}
\begin{tabular}{cccc}
\toprule
\textbf{Configuration} & \textbf{Category} & \textbf{\# of IMUs} & \textbf{IMU used}  \\
\midrule
IMU-S0 & Baseline  & 1 & 0 \\
\midrule
IMU-S2 & Symmetric & 2 & 1, 2 \\
IMU-S4 & Symmetric & 4 & 1, 2, 3, 4 \\
IMU-S6 & Symmetric & 6 & 1, 2, 3, 4, 5, 6 \\
\midrule
IMU-A2 & Asymmetric & 2 & 11, 12 \\
IMU-A4 & Asymmetric & 4 & 11, 12, 13, 14 \\
IMU-S6 & Asymmetric & 6 & 11, 12, 13, 14, 15, 16 \\

\bottomrule
\end{tabular}
\vspace*{-0.2in}
\end{table}




\subsection{Results}

The results are summarized in Table \ref{tab:sim_result}. Here the rotational error is the norm of the axis-angle rotation between the estimated pose and the ground-truth pose. The positional error is the distance between the estimated position and the ground-truth position. As more and more IMUs are fused, the mean errors decrease. This matches the expectation, as the combined noises from the fused IMUs are attenuated, as shown by (\ref{noise_reduction_asym}) and (\ref{noise_reduction_sym}). Comparing asymmetric configurations against symmetric configurations, the errors are marginally larger. This can be explained by the unequal weighting, which introduces larger noises than equal weighting, given that all IMUs are identical. Overall, the results suggest that averaging between IMUs works for both gyroscope and accelerometer measurements. The combined measurements are also more accurate.

\begin{table}[h!]
\vspace*{3pt}
\centering
\caption{Simulation Results}
\label{tab:sim_result}
\begin{tabular}{ccccccc}
\toprule
\textbf{Configuration} & \multicolumn{2}{c}{\textbf{Rotational Err (rad)}} & \multicolumn{2}{c}{\textbf{Positional Err (m)}} \\
\cmidrule(lr){2-3} \cmidrule(lr){4-5}
& MAE & RMSE & MAE & RMSE \\
\midrule
IMU-S0 & 0.01568 & 0.01873 & 0.11189 & 0.13260 \\
\midrule
IMU-S2 & 0.01278 & 0.01486 & 0.10029 & 0.11540 \\
IMU-S4 & 0.01122 & 0.01319 & 0.09223 & 0.10598 \\
IMU-S6 & 0.01028 & 0.01176 & 0.08950  & 0.10271 \\
\midrule
IMU-A2 & 0.01306 & 0.01514 & 0.10055 & 0.11685 \\
IMU-A4 & 0.01129 & 0.01285 & 0.09222 & 0.10554 \\
IMU-A6 & 0.01083 & 0.01221 & 0.08944 & 0.10077 \\
\bottomrule
\end{tabular}
\end{table}

\section{Real-World Experiment}\label{real_experiment}

For real-world experiments, we chose to perform visual-inertial odometry to understand the performance of averaging in a realistic scenario.





\begin{figure*}[ht]
\vspace*{3pt}
    \centering
    \hspace{-0.4in}
    \begin{subfigure}[b]{0.45\linewidth}
        \centering
        \includegraphics[width=8.25cm]{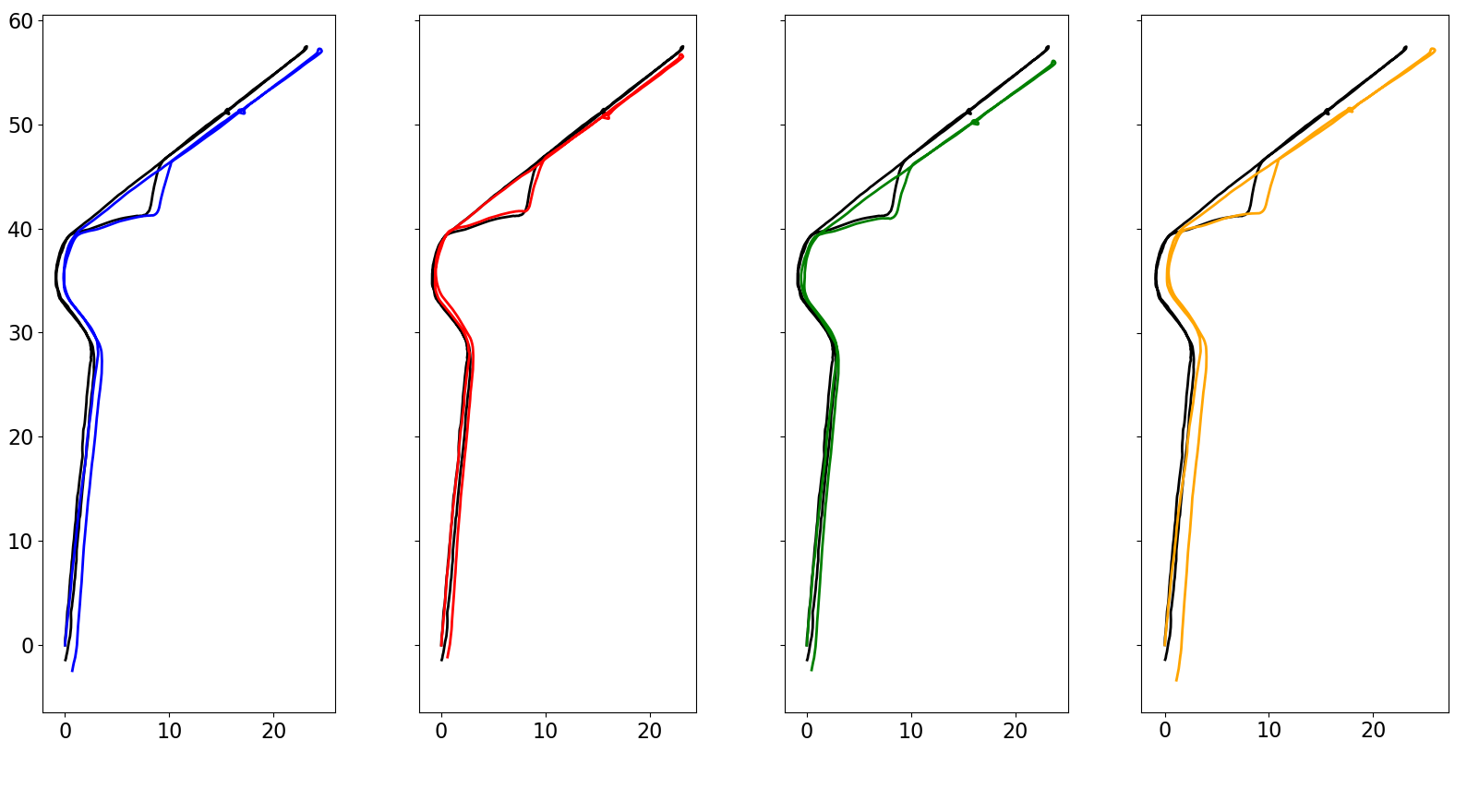}
        \caption{Example Trajectories. All trajectories start at (0, 0).}
        \label{fig:posyaw_af}
    \end{subfigure}
    \hspace{0.2in}
    \begin{subfigure}[b]{0.45\linewidth}
        \centering
        \includegraphics[width=9cm]{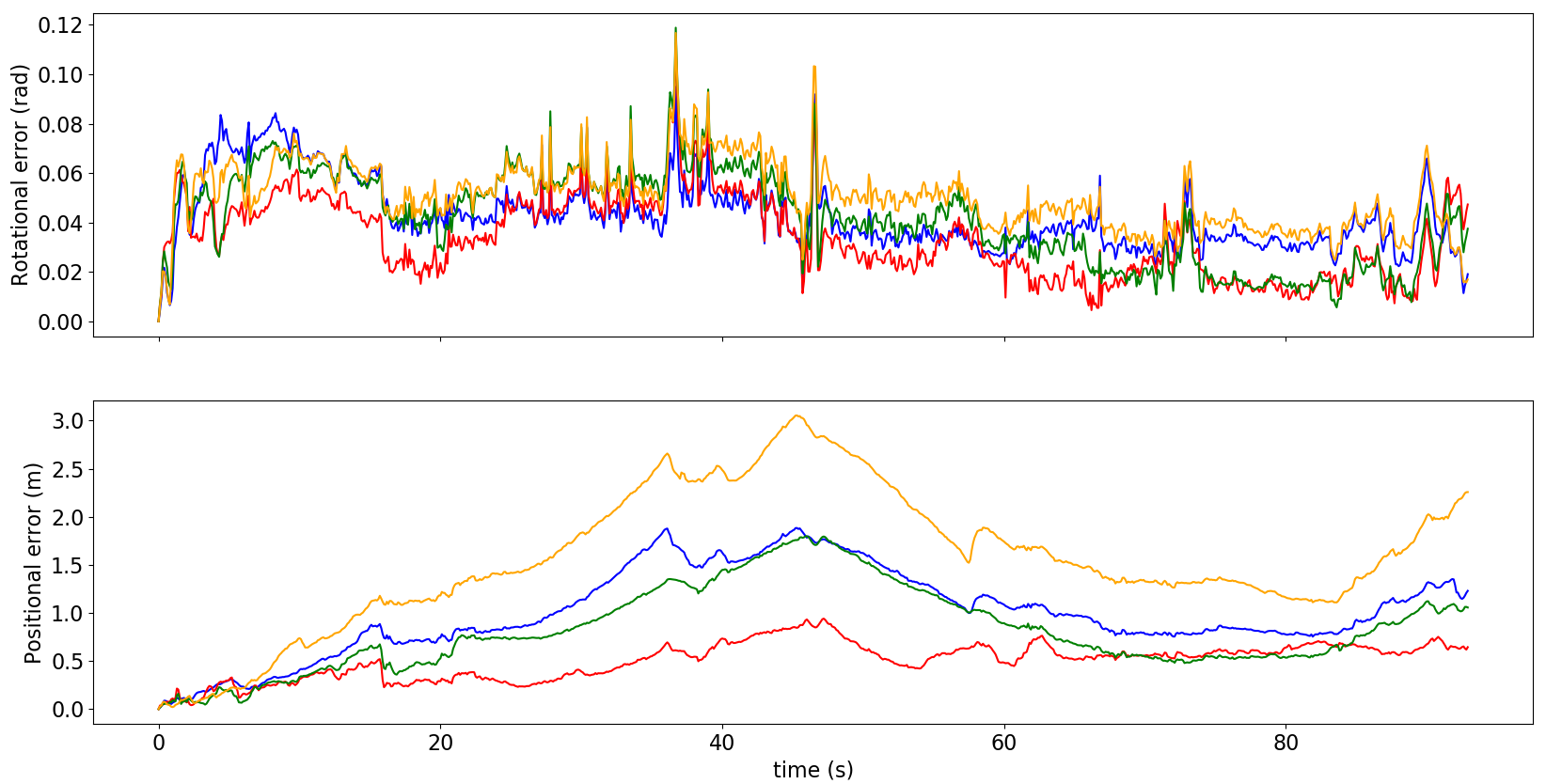}
        \caption{Example errors over Time Rot. (Top), Pos. (Bottom).}
        \label{fig:error_af}
    \end{subfigure}
\caption{Example Results for Run AF. Single (blue) Averaged (red), Centered (green), Negative (yellow). We see that averaging multiple IMUs (red/green) has less error than using a single IMU (blue).}
\label{fig:real_example}
\vspace*{-0.2in}
\end{figure*}

\subsection{Dataset}

\begin{figure}[h]
\centering
\begin{subfigure}{0.35\textwidth}
\centering
\includegraphics[width = \textwidth]{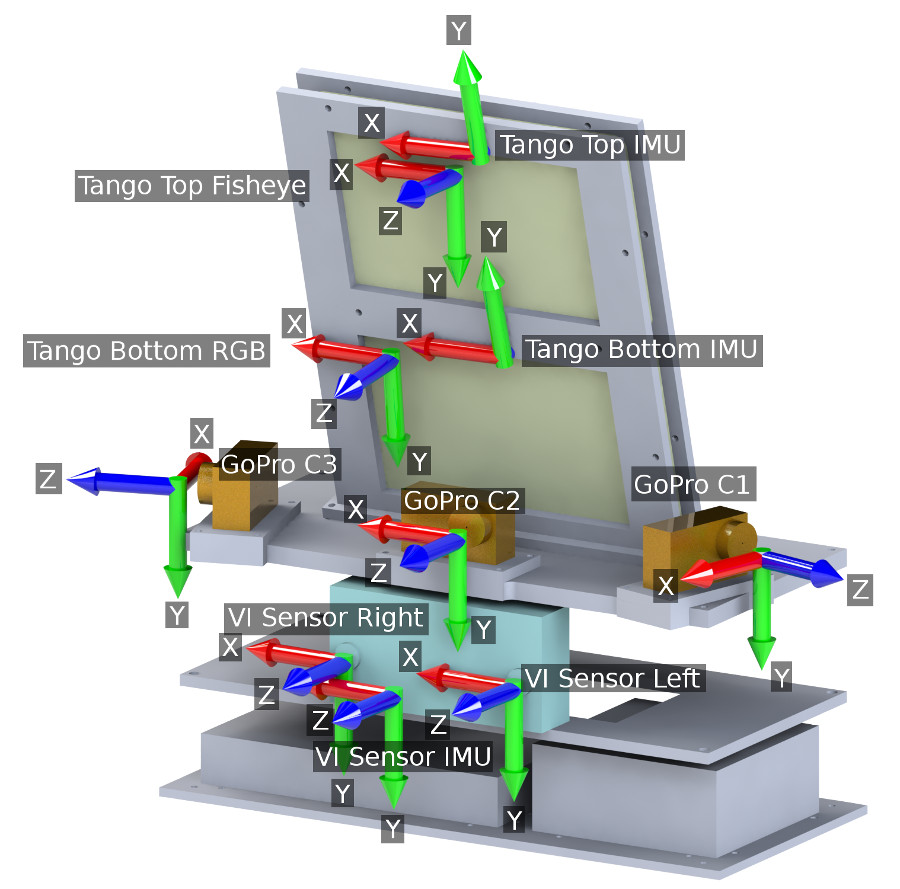}
\caption{Sensors}
\label{fig:sensor_on_rig}
\end{subfigure}
\begin{subfigure}{0.12\textwidth}
\centering
\includegraphics[width = \textwidth]{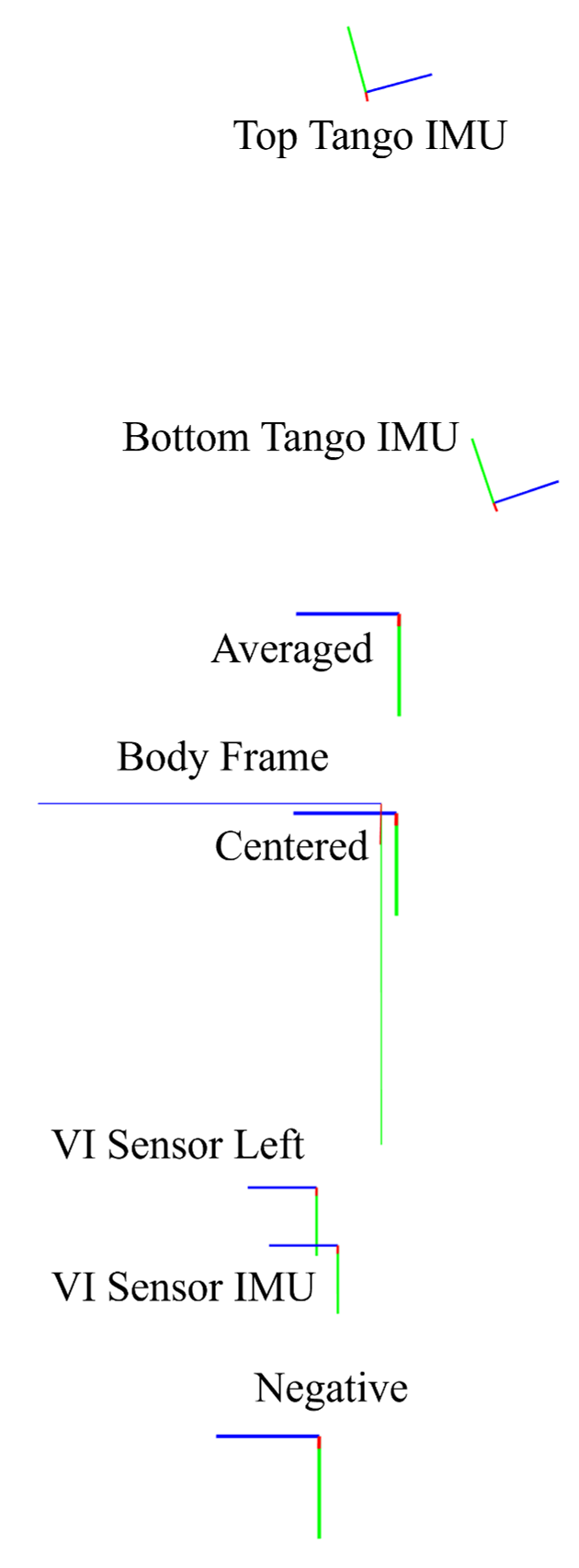}
\caption{Ref. Frames}
\label{fig:rig_coordinate_frame}
\end{subfigure}
\caption{(a) shows the type and location of sensors on the test rig (image credit: \cite{penncosyvio}) (b) shows the various frames of reference used in the experiments, viewed from the side.}
\label{fig:combined}
\end{figure}

Most public datasets only contain a single IMU. We found that the PennCOSYVIO dataset \cite{penncosyvio} met our needs of containing data from multiple IMUs. This dataset contains four different runs (AF, AS, BF, BS; A - training, B - testing, F - fast, S - slow) with sensor data from three IMUs and multiple cameras as shown in Figure \ref{fig:sensor_on_rig}. For our experiment, we chose to use only the VI sensor camera, as complete calibration information was provided. We also used only the left camera of the VI sensor, which forms the minimal configuration with a single IMU. This is to better expose the impact of different IMU configurations on performance. We compared the performance of using only the VI sensor IMU versus the VIMU from averaging all three IMUs. We selected the VI sensor IMU as the single IMU benchmark as it is more accurate compared to the IMUs in the two Tango devices.

\subsection{Implementation}

For VIO implementation, we selected OpenVINS \cite{openvins} to test the different IMU configurations. OpenVINS is based on the MSCKF \cite{Anastasios2007_MSCKF} and supports a single IMU with multiple cameras. Besides the vehicle state, it can also estimate cameras and IMU intrinsics, as well as the time offset and extrinsics between the IMU and cameras. After some tuning, we found that enabling extrinsic and time-offset estimation gave the best result over many repeated runs. This helped eliminate the effect of calibration errors in the dataset. Also, since the PennCOSYVIO dataset does not have an initialization phase when the vehicle is at rest, we enabled dynamic initialization in OpenVINS to estimate initial velocity and biases in motion.

\subsection{Setup}

To demonstrate that the weighting works as expected, we designed four configurations to test: 1) single IMU with weights $[\begin{matrix} 1.0 & 0.0 & 0.0 \end{matrix}]$ (here the first weight is for the VI sensor IMU, the second weight for the top Tango IMU, and the third weight is for the bottom Tango IMU), 2) averaged VIMU with weights $[\begin{matrix} 0.33 & 0.33 & 0.33 \end{matrix}]$, 3) VIMU closest to GoPro 2 (the defined origin of the test rig) with weights $[\begin{matrix} 0.4944 & 0.1546 & 0.3509 \end{matrix}]$ (this places the VIMU at $[\begin{matrix} 0.0186 & -0.0012 & -0.0046 \end{matrix}]$ in the frame of GoPro 2), and 4) we also designed a case where some of the weights go negative $[\begin{matrix} 1.2 & -0.1 & -0.1 \end{matrix}]$ (this places the VIMU outside of the triangle formed by the three IMUs). The various reference frames are as shown in Figure \ref{fig:rig_coordinate_frame}.

\subsection{Results}

The results are summarized in Table \ref{tab:vio_result}. In Figures \ref{fig:posyaw_af} and Figure \ref{fig:error_af}, we show the trajectories and errors for run AF as an example.
\begin{table}[h!]
\centering
\caption{PennCOSYVIO Results}
\label{tab:vio_result}
\begin{tabular}{cccccc}
\toprule
\textbf{Run} & \textbf{Config.} & \multicolumn{2}{c}{\textbf{Rotational Err (rad)}} & \multicolumn{2}{c}{\textbf{Positional Err (m)}} \\
\cmidrule(lr){3-4} \cmidrule(lr){5-6}
 & & MAE & RMSE & MAE & RMSE \\
\midrule
AF & Single    & 0.0428 & 0.0451 & 1.0031 & 1.0946 \\
AF & Averaged  & \textbf{0.0337} & \textbf{0.0373} & \textbf{0.5051} & \textbf{0.5393} \\
AF & Centered  & 0.0419 & 0.0459 & 0.8258 & 0.9331 \\
AF & Negative  & 0.0505 & 0.0523 & 1.5588 & 1.7128 \\
\midrule
AS & Single    & 0.0400 & 0.0433 & 1.4822 & 1.5485 \\
AS & Averaged  & 0.0421 & 0.0462 & 1.4427 & 1.4902 \\
AS & Centered  & \textbf{0.0324} & \textbf{0.0365} & \textbf{1.3765} & \textbf{1.4387} \\
AS & Negative  & 0.0759 & 0.0821 & 1.7956 & 2.0234 \\
\midrule
BF & Single    & 0.0890 & 0.2983 & 1.1189 & 1.2289 \\
BF & Averaged  & 0.0794 & \textbf{0.2967} & \textbf{0.4917} & \textbf{0.5262} \\
BF & Centered  & \textbf{0.0787} & 0.2967 & 0.7021 & 0.7417 \\
BF & Negative  & 0.1082 & 0.3020 & 3.4679 & 3.6277 \\
\midrule
BS & Single    & 0.0402 & 0.0659 & 1.5926 & 1.7065 \\
BS & Averaged  & 0.0353 & 0.0635 & 2.1734 & 2.2367 \\
BS & Centered  & \textbf{0.0290} & \textbf{0.0592} & \textbf{1.0496} & \textbf{1.1091} \\
BS & Negative  & 0.0490 & 0.0714 & 1.8005 & 1.9496 \\
\bottomrule
\end{tabular}
\vspace*{-0.1in}
\end{table}
From the results, we can clearly see that the averaged and centered configurations work better than using a single IMU. This is expected, as averaging can reduce noise and slow down bias drift, thus resulting in more accurate trajectory tracking. We also see that the negative configuration has larger error over time. This is due to the fact that when the norm of the weight vector is larger than 1, noise is amplified.

\section{Conclusion and Future Work}

In this paper, we presented new way to average individual IMUs into a single virtual IMU. This makes our methods compatible with existing single-IMU estimators. Our VIMU has no lever-arm terms in its propagation model even when the individual IMUs are far apart.  We provided a closed-form solution for determining the weights to place the VIMU frame at a preferred location (e.g., same as a camera frame) while also minimizing measurement noise. Through both simulation and a real dataset experiment, we validated the main concept of our VIMU frame placement and demonstrated the noise-reduction benefit of averaging.

For future work we are interested in incorporating online calibration and fault detection. We believe fault detection could be accomplished by looking at the residual errors in the IMU averaging and then appropriately down-weighting outliers, while keeping the VIMU in the same location.

\section{Acknowledgements}

The authors would like to thank the Natural Sciences and Engineering Research Council (NSERC) of Canada for supporting this work.

\newpage
\bibliography{References.bib}

\end{document}